\documentclass[conference]{IEEEtran}

\usepackage{algorithm}
\usepackage{algorithmicx}
\usepackage{algpseudocode}
\usepackage{amsfonts}
\usepackage{amsmath}
\usepackage{amssymb}
\usepackage[ansinew]{inputenc} 
\usepackage{xcolor}
\usepackage{mathtools}
\usepackage{graphicx}
\usepackage{caption}
\usepackage{subcaption}
\usepackage{import}
\usepackage{multirow}
\usepackage{cite}
\usepackage[export]{adjustbox}
\usepackage{breqn}
\usepackage{mathrsfs}
\usepackage{acronym}
\usepackage{longtable}
\usepackage{url}
\usepackage[english]{babel}
\usepackage{booktabs}
\usepackage{paralist}
\usepackage[justification=raggedright,singlelinecheck=false,size=footnotesize]{caption}
\usepackage[justification=raggedright,singlelinecheck=false,size=footnotesize]{subcaption}

\usepackage{setspace}
\usepackage{bm}
\usepackage{stackengine}
\usepackage{hyperref}

\usepackage{listings}

\lstdefinelanguage{BibTeX}
  {keywords={%
      @article,@book,@collectedbook,@conference,@electronic,@ieeetranbstctl,%
      @inbook,@incollectedbook,@incollection,@injournal,@inproceedings,%
      @manual,@mastersthesis,@misc,@patent,@periodical,@phdthesis,@preamble,%
      @proceedings,@standard,@string,@techreport,@unpublished%
      },
   comment=[l][\itshape]{@comment},
   sensitive=false,
  }

\usepackage{listings}

\DeclareMathOperator*{\argmax}{arg\,max}

\graphicspath{{./figures/}}
\newcommand{\fig}[1]{Fig.~\ref{#1}}
\newcommand{\tab}[1]{Tab.~\ref{#1}}

\newcommand{\mytexttilde}{{\raise.17ex\hbox{$\scriptstyle\mathtt{\sim}$}}}

\usepackage{array}
\usepackage{enumitem}
\usepackage{enumerate}
\setlist[description]{leftmargin=.5cm}

\graphicspath{{Template/images}}
\DeclareGraphicsExtensions{.pdf,.jpeg,.png,.jpg}

\begin{document}

\title{LightSNN: Lightweight Architecture Search for Sparse and Accurate Spiking Neural Networks
\thanks{This work has been supported by the EU H2020 MSCA ITN project Greenedge (grant no. 953775), by the EU through the Horizon Europe/JU SNS project ROBUST-6G (grant no. 101139068), and by the EU under the Italian National Recovery and Resilience Plan (NRRP) of NextGenerationEU, partnership on ``Telecommunications of the Future'' (PE0000001 - program ``RESTART'').}

}

\author{\IEEEauthorblockN{Yesmine Abdennadher$^*$, Giovanni Perin$^{\dag,*}$, Riccardo Mazzieri$^*$, Jacopo Pegoraro$^*$, and Michele Rossi$^*$}
\IEEEauthorblockA{$^*$Department of Information Engineering (DEI), University of Padova, Padova, Italy}
\IEEEauthorblockA{$^\dag$Department of Information Engineering (DII), University of Brescia, Brescia, Italy}
}

\IEEEoverridecommandlockouts

\newcounter{remark}[section]
\newenvironment{remark}[1][]{\refstepcounter{remark}\par\medskip
   \textbf{Remark~\thesection.\theremark. #1} \rmfamily}{\medskip}

\maketitle

\begin{abstract}
Spiking Neural Networks (SNNs) are highly regarded for their energy efficiency, inherent activation sparsity, and suitability for real-time processing in edge devices.
However, most current SNN methods adopt architectures resembling traditional artificial neural networks (ANNs), leading to suboptimal performance when applied to SNNs. While SNNs excel in energy efficiency, they have been associated with lower accuracy levels than traditional ANNs when utilizing conventional architectures. In response, in this work we present LightSNN, a rapid and efficient Neural Network Architecture Search (NAS) technique specifically tailored for SNNs that autonomously leverages the most suitable architecture, striking a good balance between accuracy and efficiency by enforcing sparsity. Based on the spiking NAS network (SNASNet) framework, a cell-based search space including backward connections is utilized to build our training-free pruning-based NAS mechanism. Our technique assesses diverse spike activation patterns across different data samples using a sparsity-aware Hamming distance fitness evaluation. Thorough experiments are conducted on both static (CIFAR10 and CIFAR100) and neuromorphic datasets (DVS128-Gesture). Our LightSNN model achieves state-of-the-art results on CIFAR10 and CIFAR100, improves performance on DVS128Gesture by 4.49\%, and significantly reduces search time most notably offering a $98\times$ speedup over SNASNet and running 30\% faster than the best existing method on DVS128Gesture.
Code is available on Github at: \href{https://github.com/YesmineAbdennadher/LightSNN}{https://github.com/YesmineAbdennadher/LightSNN}.
\end{abstract}



\section{Introduction}
\label{sec:introduction}

After the development of perceptrons and artificial neural networks (ANNs)~\cite{Goodfellow-et-al-2016}, spiking neural networks (SNNs)~\cite{maass1997networks}~\cite{Gerstner_Kistler_2002} emerge as the third generation of neural networks. SNNs mimic the behavior of biological neurons through discrete and time-dependent signals known as {\it spikes}. This makes them suitable for temporal (1D) and spatiotemporal (3D) data processing, offering better efficiency and reduced energy consumption compared to conventional neural networks. Because of the event-driven and low-power nature of SNNs, they have attracted major attention in the fields of edge computing, wearable devices, and signal processing at the physical layer of wireless systems. In fact, SNNs are eminently suitable for use in battery-powered or resource-limited devices due to their low energy consumption and real-time adaptation capabilities. Thanks to their amenability to on-chip implementation~\cite{8259423}~\cite{7229264} and low-latency handling of signals, they bear the promise to suit those applications where efficiency and speed are of the essence, such as wireless communications~\cite{10472876}~\cite{9771543} and biomedical signal processing ~\cite{10.3389/fnins.2021.651762}~\cite{kim2024exploring}, among others.

Previous work has focused on developing learning algorithms and training protocols; for example, based on spike-timing-dependent plasticity (STDP) or event-driven methods, etc.~\cite{guo2021neural,rathi2021diet,wang2022ltmd,bu2023optimal,neftci2019surrogate,bellec2018long,fang2021incorporating}. However, minor attention has so far been paid to architectural designs, preventing SNNs from fully leveraging their characteristics. Consequently, SNNs are still behind ANNs in terms of scalability to deep network models and performance (especially accuracy). To address these performance gaps and strike a balance in the trade-off between accuracy and energy efficiency, novel architectural designs are to be explored.

\begin{figure*}[tb]
    \centering
  \includegraphics[width=0.85\textwidth]{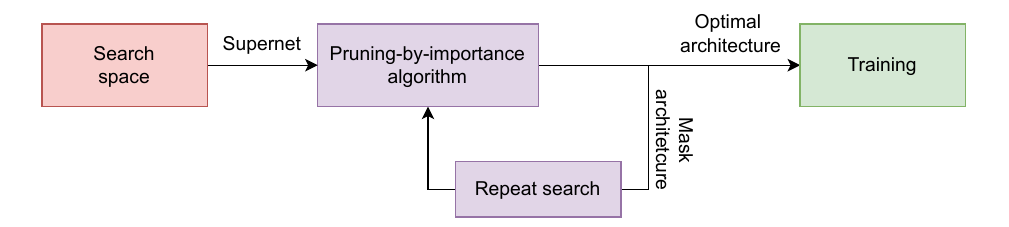}
  \caption{High-level diagram of the proposed NAS framework.}
  \label{fig:pipeline}
\end{figure*}

Neural architecture search (NAS)~\cite{ren2021comprehensive} has propelled AI forward by automatically exploring the design space to identify high-performance architectures, minimizing manual tuning and producing highly efficient models. In the past few years, NAS has found very effective neural network architectures that have succeeded in several tasks, including image segmentation~\cite{zhang2021dcnas,weng2019unet,liu2019auto,wang2021bix}, object detection~\cite{yan2021lighttrack,chen2019detnas,guo2020hit,jiang2020sp}, and other challenging domains such as speech~\cite{kim2020evolved} and image recognition~\cite{chen2021glit,zhou2022training}. With NAS, a systematic search for optimal architectural designs is performed. In doing so, the event-driven nature of SNNs can be exploited, and the need for manual experimentation can be substantially reduced. Through a methodical assessment of various network topologies, NAS can identify configurations that effectively balance energy efficiency and accuracy, by tailoring the search strategy to the specific dynamics of SNNs.

NAS techniques have recently been investigated for SNNs. However, previous works ignored the effectiveness (e.g., computation cost) of the search methods used. For example, the techniques presented in~\cite{yan2024efficient} and~\cite{na2022autosnn} involve training a supernet prior to the actual search stage which can be very expensive in terms of both GPU hours and memory usage, while~\cite{pan2024brain} uses a computationally demanding performance predictor, which requires prior training on a small subset of potential architectures. During this initial training phase, performance indicators such as early accuracy are collected are used to train a regression model. Afterwards, the trained predictor infers how unseen architectures might perform without needing to train these all the way. The approach in~\cite{li2020autost} shows good accuracy performance, but its main drawback is that this comes at the cost of a high number of floating-point operations (FLOPS), which goes against our efficiency requirement. SNASNet~\cite{kim2022neural} is the first algorithm to use a \emph{training-free} NAS technique for SNNs, thus enabling a more efficient search phase. SpikeNas, the recently proposed optimizations technique of~\cite{putra2024spikenas} modify the original SNASNet framework to further reduce the search space, while at the same time coping with memory constraints. To the best of our knowledge, this is to date the most effective algorithm in terms of search time, accuracy and sparsity of the found network models. Our solution will achieve notable improvements, especially in dynamic datasets.  

To devise our proposed technique, LightSNN, we first analyze SNASNet and highlight its limitations. Next, we delve into our enhancements, outlining how we modified the framework to get around its shortcomings and obtaining improved results. In Fig.~\ref{fig:pipeline}, we show an overview of the general phases that drive the proposed NAS algorithm. The main objectives of our design are as follows:

\begin{itemize}
    \item \textbf{Improving model accuracy.} By effectively searching across the entire search space, to evaluate the importance of each operation, we achieve new state-of-the-art accuracies on static datasets, and a substantial $4.94$\% accuracy improvement on an event-based dataset. 
    \item \textbf{Reducing network complexity.} The search space has been reduced by eliminating those benchmark operations that lead to a minor improvement in the network task performance.
    \item \textbf{Enforcing sparsity in the final architecture.} Using various operations, such as zeroize and max-pooling, allowed us to reduce the sparsity of the final model that is outputted by our NAS algorithm.
\end{itemize}

The paper is organized as follows. Section~\ref{sec:related_work} briefly reviews the baseline frameworks and methods that were considered as a starting point for the design of LightSNN, our newly proposed NAS algorithm. LightSNN is presented in Section~\ref{sec:processing_architecture} alongside its design principles. The results for the new NAS framework are reported in Section~\ref{sec:numerical_results} for both static and dynamic datasets. Our final considerations are drawn in Section~\ref{sec:conclusions}.


\section{Baseline approaches}
\label{sec:related_work}

\subsection{Spiking neuron dynamics}
In contrast to traditional artificial neurons, which accumulate real-valued inputs and apply a non-linear activation function (such as ReLU) to produce real-valued outputs, spiking neurons operate differently, by mimicking the fundamental behavior of biological neurons. These computational units aggregate inputs over a number of timesteps within their membrane potential, and an output spike is generated only when the membrane potential reaches a predefined threshold. This characteristic spiking behavior is captured by the leaky integrate and fire (LIF) neuron model~\cite{brunel2007lapicque}, where the neuron's membrane potential $v(t)$ increases with each incoming spike, but experiences a ``leakage'' with time $t$, causing the membrane potential to decay. Upon reaching a threshold $V_{\rm th}$, the neuron fires an output spike and the membrane potential is reset to a resting value $V_{\rm reset}$.

\subsection{SNASNet}

SNASNet~\cite{kim2022neural} employs iterative search within a subset of architectures chosen from a larger pool of more than $200$ million candidates randomly generated from a cell-based search space. It assesses \emph{a priori} each architecture using a {\it training-free} metric to determine which architecture could obtain the highest accuracy after training.

\subsubsection{Cell-based search space}

The search space comprises a macro skeleton and a micro skeleton, the first includes a stem layer consisting of a convolution layer that extracts the first feature maps, two identical cells to be searched, a reduction cell, and a classifier.

\begin{figure}[tb]
  \centering
  \includegraphics[width=\columnwidth]{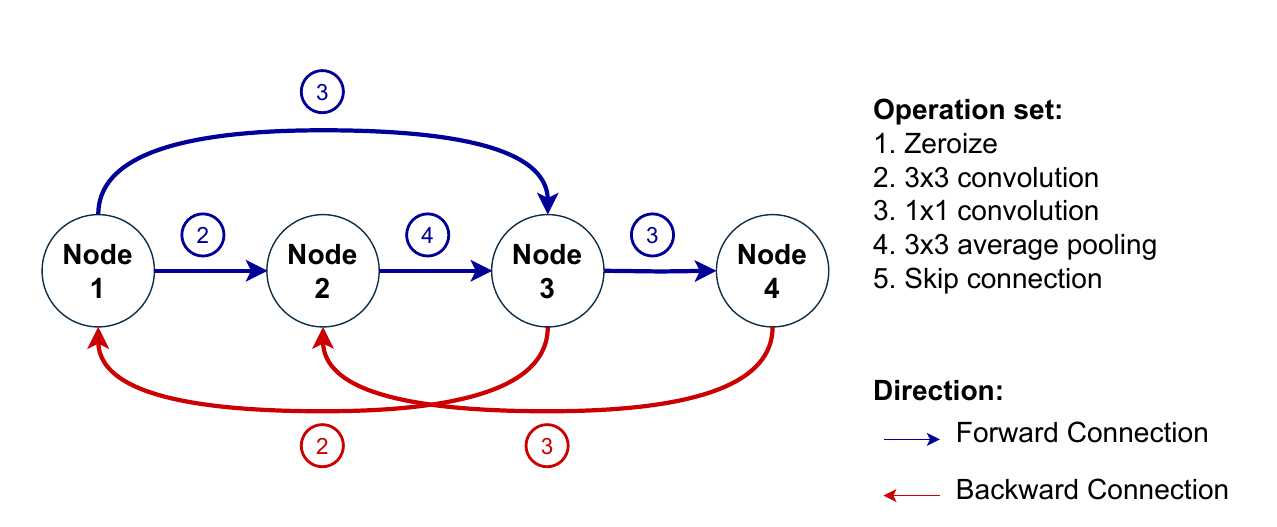}
  \caption{Example of a candidate cell.}
  \label{fig:cell}
\end{figure}

Each searched cell consists of a micro skeleton including four nodes, with each node capable of connecting to another via an operation from the set $\mathcal{O} = \{$\textit{zeroize, skip connection, 1x1 convolution, 3x3 convolution, 3x3 average pooling}$\}$ as shown in the example of a candidate cell in Fig.~\ref{fig:cell}. The Zeroize operation is designed to force selected elements to zero. Each node within the network keeps the cumulative value of the arriving feature maps after being applied to existing operations. Backward connections are included too. The backward operation involves adding a transformed node feature from the $l$-th layer at timestep $t-1$ to the node of the $l'$-th layer at timestep $t$ (where $l' < l$). The backward connections also adhere to the same operation set search space $\mathcal{O}$ as the forward connections. 

\subsubsection{Training-free NAS approach}
The sparsity-aware Hamming distance (SAHD) evaluates architectures based on their performance at initialization without requiring training. Architectures that generate distinct representations across different samples are likely to achieve high accuracy after training~\cite{kim2022neural}. Specifically, the SAHD measures the difference between binary codes (activation patterns) produced by the untrained network for input data pairs within a mini-batch. A greater distance between these activation patterns suggests a higher post-training accuracy. To analyze the relation between binary codes for an entire mini-batch of size $N$, we calculate the kernel matrix $K^{(t)}_{H}$ as
\begin{equation}
K^{(t)}_{H} =
\begin{pmatrix}
N_A - d^{(t)}(c_1, c_1)  & \cdots & N_A - d^{(t)}(c_1, c_N) \\
\vdots & \ddots & \vdots \\
N_A - d^{(t)}(c_N, c_1)  & \cdots & N_A - d^{(t)}(c_N, c_N) \\
\end{pmatrix},
\label{kh}
\end{equation}
where $N_A$ is the number of LIF neurons and $d(c_i, c_j)$ represents the SAHD between binary activations $c_i$ and $c_j$ for data samples $i$ and $j$.
The global SAHD score is computed by accumulating the SAHD across all layers, and it is used to generate the kernel matrix \eqref{kh} at each timestep $t$. Next, we use the following equation to aggregate the kernel matrices and determine the final score $s$~\cite{kim2022neural}

\begin{equation}
s = \log \left[\det \left( \left\lvert \sum_{t} K^{(t)}_{H} \right\rvert \right) \right] .
\label{s}
\end{equation}

\section{LightSNN: rationale and methods}
\label{sec:processing_architecture}

\subsection{Pruning-by-importance algorithm}

The iterative random search method basically relies on chance-driven selection over only a small subset of the entire search space ($5,000$ candidates) and, thus, may result in many suboptimal architectures, which increases the chance of missing other promising designs in the more extensive search space. Furthermore, the individual evaluation of each architecture can be time-consuming and computationally intensive. A faster and more reliable search technique is required to circumvent these restrictions and increase the likelihood of finding an architecture with good performance. The application of the \emph{pruning-by-importance} algorithm, as shown in~\cite{chen2021neural}, is a good strategy to achieve this goal. To the best of our knowledge, our work is the first to apply this approach to NAS for SNNs.

While exploring the search space, possible architecture candidates in each cell have $E$ edges connecting the nodes and various operators in the previously defined set $\mathcal{O}$, of cardinality $O$. Sampling methods require examining $O^E$ unique cells, leading to a search complexity of $\Theta\left(O^E\right)$. For SNASNet, this means that we would need to evaluate $5^{12} \approx 2.4 \times 10^8$ possible architectures (including backward connections). The pruning-by-importance algorithm~\cite{chen2021neural}, in contrast, takes a different approach by assessing a supernet that includes every possible operator and edge. This approach significantly reduces the search complexity, by boosting effectiveness. In fact, the exploration cost is lowered from $\Theta\left(O^E\right)$ to $\Theta\left(O \cdot E\right)$, providing a more effective and economical resource consumption. This means that, in our case, this strategy reduces the complexity of the evaluation to $5 \cdot 12 = 60$ iterations.

The pruning-by-importance algorithm is composed of two loops, detailed in what follows and referring to Algorithm~\ref{algo1}.

\begin{description}
    \item[Outer loop:] At each round, a single operator is pruned (removed) from each edge. This outer loop continues until the current supernet transforms into a single-path network, which represents the stopping condition. This returns the architecture identified through the search (line \ref{line10}).
    \item[Inner loop:] The significance of individual operators is evaluated by computing ${\rm SAHD}_{N\backslash o_j}$. With the notation $N\backslash o_j$ we refer to a network $N$ where the operation $o_j$ has been pruned. The operator with the least impact on the SAHD is considered the least influential (line \ref{line6}). Thus, the operation $o_j^*$ whose removal leads to the highest score ${\rm SAHD}_{N\backslash o_j^*}$ is removed from each edge (line \ref{line7}).
\end{description}

\begin{figure}[tb]
     \centering
     \begin{subfigure}{0.45\columnwidth}
         \centering
         \includegraphics[width=0.75\textwidth, trim=0.75cm 1.5cm 0.75cm 1.5cm, clip]{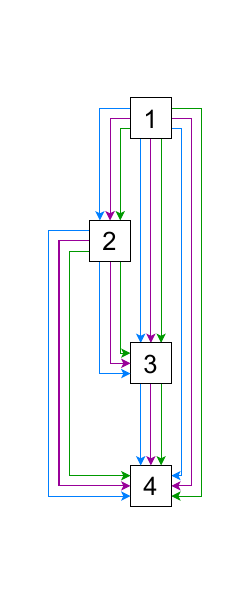}
         \caption{Supernet (initial state).}
         \label{fig:a}
     \end{subfigure}
     \hspace{0.05\columnwidth}
     \begin{subfigure}{0.45\columnwidth}
         \centering
         \includegraphics[width=0.75\textwidth, trim=0.75cm 1.5cm 0.75cm 1.5cm, clip]{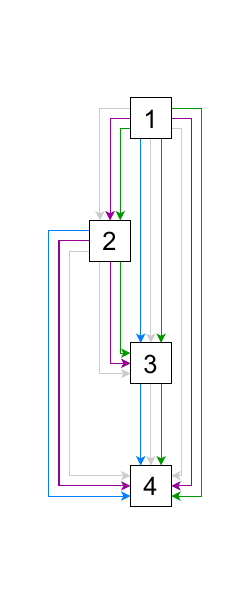}
         \caption{Prune one operation.}
         \label{fig:b}
     \end{subfigure}
     \begin{subfigure}{0.45\columnwidth}
         \centering
         \includegraphics[width=0.75\textwidth, trim=0.75cm 1.5cm 0.75cm 1.5cm, clip]{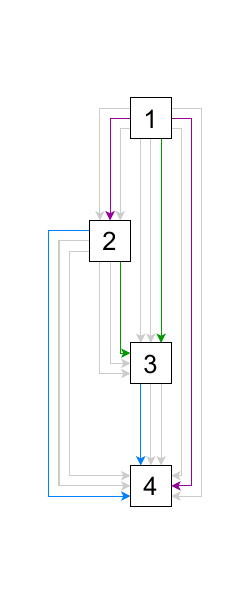}
         \caption{Prune second operation}
         \label{fig:c}
     \end{subfigure}
     \hspace{0.05\columnwidth}
     \begin{subfigure}{0.45\columnwidth}
         \centering
         \includegraphics[width=0.75\textwidth, trim=0.75cm 1.5cm 0.75cm 1.5cm, clip]{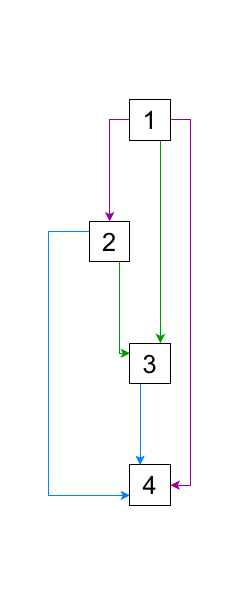}
         \caption{Single-path network}
         \label{fig:d}
     \end{subfigure}
        \caption{Pruning by importance steps. Pruned operations are shown with a light-gray color.}
        \label{fig:4grph}
\end{figure}

In \fig{fig:4grph}, we present a simplified representation of the pruning process within a cell consisting of forward connections, each with three possible operations. Each operation is denoted by a distinct color. The process unfolds as follows:
\begin{itemize}
    \item In the initial state, all operations are considered ($O$ operations are available at each node, see \fig{fig:a}).
    \item The importance of each operation is evaluated and the least significant ones are pruned (\fig{fig:b}).
    \item The process is iterated, eliminating the second least important operations (\fig{fig:c}).
    \item Finally, a single-path network is obtained, retaining only the most crucial operations (\fig{fig:d}).
\end{itemize}

\begin{algorithm}[tb]
\caption{Operator Pruning Algorithm}
\label{algo1}
\textbf{Input:} Supernet $N$ stacked by cells, each cell with $E$ edges, each edge with $O$ operators.
\begin{algorithmic}[1]
\While{$N$ is not a single-path network}
    \For{each operator $o_j$ in $N$}
        \State $ s_{N\backslash o_j} \gets {\rm SAHD}_{N\backslash o_j} $ \Comment The higher $s_{N_t\backslash o_j}$ the more likely we prune $o_j$     \label{line:loop1}
    \EndFor    
    \For{each edge $e_i$, $i = 1, \ldots, E$}
        \State $j^* \gets \argmax_j\{s_{N\backslash o_j} : o_j \in e_i\}$ \label{line6}
        \State $N \gets N\backslash o_{j^*}$  \label{line7}  
    \EndFor
\EndWhile
\State \textbf{return} Pruned single-path network $N$ \label{line10}
\end{algorithmic}
\end{algorithm}

We conducted a comparison on the SNASNet search space using the original search algorithm and then the proposed pruning algorithm, on the CIFAR10 dataset. The statistics in Table~\tab{tab1} show the promising results of the pruning-by-importance in both accuracy and search time. While the random selection of a small set of candidates leaves some regions unturned, our greedy approach can navigate through a larger portion of the search space. As a consequence, using the pruning-by-importance algorithm not only reaches better performance in accuracy ($92.59$\% compared to $91.83$\% of the random search) but achieves a search time of only $2$ hours and $16$ minutes unlike the $2$ hours and $49$ minutes of the random search method (i.e., it is $20$\% faster). Thus, the efficiency of the algorithm has drastically improved. Similar results were consistently obtained throughout our experiments, providing empirical evidence that pruning by importance can better cover the search space and find good architectures within a shorter time span. 

\begin{table}[tb]
\centering
\small{
\resizebox{0.9\columnwidth}{!}{%
\begin{tabular}{@{}ccc@{}}
\toprule
\textbf{Search algorithm} & \textbf{Accuracy} & \textbf{Search time}      \\ \midrule
Random search   &   $91.83\%$       &    $2$h $49$min     \\
Pruning-by-importance &    $92.59\%$       &  $2$h $16$min \\
\bottomrule
\end{tabular}%
}}
\caption{Search algorithms comparison.}
\label{tab1}
\end{table}
\subsection{Lightweight-and-sparsity-aware NAS}

\subsubsection{Search space refinements}

It has been noted that a sparser SNN provides better efficiency and performance~\cite{yao2023sparser}. Sparsity consists in the reduction of the number of active connections and neurons over time, which leads to enhanced energy efficiency. This also allows for a lower memory footprint and, hence, more efficient hardware utilization. Additionally, sparsity improves the network generalization capabilities thanks to its inherent regularization effect~\cite{Muthukumar2023SparsityawareGT}.
To leverage a sparser spiking neural network, we perform some design modifications to the search space, creating a more efficient and lightweight framework via the following expedients.

\begin{description}
    \item[Max pooling:]
Replacing average pooling with max pooling in SNNs preserves the binary spiking behavior by selecting the most significant spike within a pooling window, ensuring that only the most critical spikes are propagated. Hence, this approach maintains the binary nature of SNNs while promoting sparsity, leading to energy efficiency by reducing the number of spikes~\cite{na2022autosnn}. 

    \item[3-operation-cell:]
Building on the analytical comparison of operations conducted in~\cite{putra2024spikenas} to quantify the importance of each operation, we opted to reduce the number of operations in our search space to three. The ranking conducted in~\cite{putra2024spikenas} prioritized operations as follows: 
\begin{inparaenum}[(1)]
    \item $3\times3$ convolution,
    \item skip connection, and
    \item zeroize, $1\times 1$ convolution, and $3\times3$ average pooling have the same importance.
\end{inparaenum}
They selected $3\times3$ convolution, skip connection, and average pooling for their architecture search space. Thus, for our own solution, we choose to work with $3\times3$ convolution,
skip connection, and in contrast, we replace average pooling with zeroize in our design to enhance network sparsity. 
\end{description}

\section{Experiments and evaluation}
\label{sec:numerical_results}

\begin{table*}[t]
\centering
\resizebox{0.9\textwidth}{!}{%
\begin{tabular}{@{}ccccccc@{}}
\toprule
\textbf{Dataset} & \textbf{Method} & \textbf{Search space structure} & \textbf{Timesteps} & \textbf{Accuracy}  & \textbf{Search time} & \textbf{SAR}\\
\midrule
    &    SNASNet \cite{kim2022neural} &2 cells 5 operations&5 & $91.83\%$ &2h 49min & 0.12\\
       
 CIFAR10 &  SpikeNas\cite{putra2024spikenas} &2 cells 2 operations&5&  $\mathbf{93.18\%}$&\textbf{29s} & \textbf{0.08}\\  
      &  LightSNN (ours)&2 cells 3 operations&5&  $93\%$ & 2min 44s & 0.09\\
\midrule
        
         & SNASNet\cite{kim2022neural} &2 cells 5 operations& 5 & $\mathbf{72.36\%}$ & 2h 2min & \textbf{0.12}\\
CIFAR100 &  SpikeNas\cite{putra2024spikenas} &2 cells 3 operations&5&  $45.77\%$ &3min 3s  & \textbf{0.12}\\  
         & LightSNN (ours) &2 cells 3 operations& 5 & $70.44\%$ &  \textbf{5min 58s} &  0.13\\
\midrule
         &  SNASNet \cite{kim2022neural} & 2 cells 5 operations &16& $89.93\%$ & 11h 29min& 0.13\\
DVS128Gesture      &  SpikeNas\cite{putra2024spikenas} &2 cells 3 operations&16& $87.84\%$ & 9min 55s & \textbf{0.05}\\
  
    &   LightSNN (ours) & 2 cells 3 operations&16& $\mathbf{94.44\%}$  & \textbf{6min 54s} & 0.07\\
\bottomrule
\end{tabular}
}
\caption{Performance comparison of NAS methods for different datasets. }
\label{tab2}
\end{table*}

\subsection{Datasets}
The three popular datasets CIFAR10, CIFAR100, and DVS128 Gesture are used to assess our NAS technique. Often used as image classification benchmarks, CIFAR-10 and CIFAR-100 are static datasets consisting of $10$ and $100$ object classes, respectively, comprising low-resolution (32x32) RGB images. The event-based dataset DVS128 Gesture, which records gestures using a dynamic vision sensor (DVS), is instead used for the gesture recognition task. It includes asynchronous event streams captured from eleven dynamic hand movements. This dataset is perfectly suited for SNNs since it is conceived for neuromorphic computing and event-based processing.

\subsection{Hyperparameters}
For the search phase, weights are randomly initialized with the Kaiming Initialization~\cite{he2015delving}, and the search batch size is set to $32$. Different search batch sizes could have been investigated but were not tested in this work. For a $300$-epoch training phase, the surrogate gradient method is run to enable backpropagation in SNNs~\cite{neftci2019surrogate}, the batch size was set to $64$, with $0.2$ as the learning rate with a cosine-annealing learning rate schedule. We used the vanilla SGD optimizer with a momentum of $0.9$ and weight decay of $0.0005$. The algorithm was implemented using Pytorch and SpikingJelly libraries and executed on an Nvidia A40 GPU.

\subsection{Results}
Using the three aforementioned datasets and computing setup, we make a targeted comparative analysis of our approach, LightSNN, and state-of-the-art methods based on the SNASNet framework. Rather than comparing to all techniques using diverse frameworks, we focus on a more direct and relevant assessment within the SNASNet-based search space.

To assess the effectiveness of our work, we consider three criteria: accuracy, search time, and spiking activity rate (SAR). The accuracy informs us about how well the model could learn and generalize to test data. The search time evaluates the efficiency of the proposed NAS framework and confirms its rapidity. Finally, the SAR quantifies the frequency of spike generation within the network, directly corresponding to the energy consumption of an SNN ~\cite{yan2024reconsidering}. Specifically, the SAR is used as a proxy metric for sparsity, which generally results in lightweight and energy-efficient SNN models~\cite{yao2023sparser}. The SAR metric is here evaluated by dividing the total number of spikes by the total number of neurons multiplied by the total number of time steps. 


Compared to SNASNet, lightSNN achieves significantly higher accuracy on CIFAR-10 and DVS128Gesture and comparable accuracy on CIFAR-100, while substantially reducing search time across all datasets. It also yields lower sparsity on CIFAR-10 and DVS128Gesture and maintains similar sparsity on CIFAR-100.
Although lightSNN incurs a modest runtime penalty relative to SpikeNAS on CIFAR-10 and CIFAR100, due to its search method, it achieves substantially higher accuracy on CIFAR-100 and nearly matches SpikeNAS on CIFAR10 at equivalent sparsity. Conversely, on DVS128Gesture, lightSNN not only outperforms SpikeNAS but also runs faster while maintaining similar sparsity. 

We highlight our model's exceptional performance on the DVS128Gesture benchmark, underscoring its suitability as a prime candidate for event-based data classification.


\section{Conclusion}
\label{sec:conclusions}

In this work, we presented a new network architecture search framework to identify energy-efficient spiking neural network architectures. Building on previous research and, in particular, on the SNASNet algorithm, we propose a series of improvements with the following purposes: {\it i)} being able to explore a larger portion of solutions with respect to what SNASNet does (enlarging the search space), {\it ii)} reducing the search time for the final architecture, and {\it iii)} reducing the final model complexity in terms of number of connections and generated spikes. The findings are promising, showing a significant reduction in the time required for the architecture search to complete, alongside considerable enhancements in accuracy, complexity, and sparsity of the resulting models.

\bibliography{biblio}
\bibliographystyle{ieeetr}

\end{document}